%% file: main.tex
\documentclass[runningheads]{llncs}

\usepackage{textcomp}
\usepackage{graphicx}

\usepackage{caption}
\usepackage{subcaption}
\usepackage{comment}

\usepackage[pdftex,dvipsnames]{xcolor}  
\usepackage{hyperref} 

\newcommand{\methodname}{FoMO}
\newif\ifjournal
\newif\ifanonym

\begin{document}
        
\title{Large-Scale Video Analytics \\ through Object-Level Consolidation }
\titlerunning{Large-Scale Video Analytics through Object-Level Consolidation}

\author{Daniel Rivas\inst{1,2}\and
Francesc Guim\inst{3}\and
Jord\`a Polo\inst{1}\and
David Carrera\inst{1}}

\authorrunning{D. Rivas et al.}

\institute{Barcelona Supercomputing Center (BSC), Barcelona, Spain \and
Universitat Polit\`ecnica de Catalunya, Barcelona, Spain
\email{\{daniel.rivas,jorda.polo,david.carrera\}@bsc.es} \and
Intel Iberia, Barcelona, Spain\\
\email{francesc.guim@intel.com}}

\maketitle

\begin{abstract}
As the number of installed cameras grows, so do the compute resources required to process and analyze all the images captured by these cameras. Video analytics enables new use cases, such as smart cities or autonomous driving. At the same time, it urges service providers to install additional compute resources to cope with the demand while the strict latency requirements push compute towards the end of the network, forming a geographically distributed and heterogeneous set of compute locations, shared and resource-constrained. Such landscape (shared and distributed locations) forces us to design new techniques that can optimize and distribute work among all available locations and, ideally, make compute requirements grow sublinearly with respect to the number of cameras installed. In this paper, we present FoMO (\textit{Focus on Moving Objects}). This method effectively optimizes multi-camera deployments by preprocessing images for scenes, filtering the empty regions out, and composing regions of interest from multiple cameras into a single image that serves as input for a pre-trained object detection model. Results show that overall system performance can be increased by 8x while accuracy improves 40\% as a by-product of the methodology, all using an \textit{off-the-shelf} pre-trained model with no additional training or fine-tuning.
\end{abstract}

\input{introduction} 

\input{background}

\input{method}

\input{setup}

\input{results}

\input{related_work}

\input{conclusions}

\section*{Acknowledgments}
The authors would like to thank Dr. Josep Ll. Berral for his valuable feedback.

\bibliographystyle{splncs04}
\bibliography{references.bib}

\end{document}

%% file: introduction.tex
\section{Introduction}
\label{sed:introduction}
Rapid advances in the field of deep learning have led to an equally rapid and unprecedented increase in interest in video analytics applications. Smart cities or autonomous driving are just two examples of use cases that require an automatic analysis of video feeds to trigger different actions. However, video feeds are commonly processed individually, and, at the same time, the execution of neural networks is a computationally expensive task. This leads to the cost of video analytics systems growing linearly with the number of cameras being deployed. We believe there is a window of opportunity to optimize video analytics systems by combining mechanisms that reduce the search space within an image, thanks to our knowledge of how neural networks look for objects within the input image.

Neural networks can learn to find objects anywhere on an image, even when these objects represent a small fraction of the input and the rest of the image is of no interest. To make this possible, neural networks classify many regions (in the order of thousands) proposed at an earlier stage of the network. This implies that neural networks end up looking at regions that are empty of objects of interest. We could argue that this is what makes neural networks so powerful (especially convolutional neural networks for image analysis), as it means they can focus attention without developers or users needing to hint them where to look in the image. However, it also implies that most of the work will yield no meaningful results with an increased chance of wrong results (any positive detection on an \textit{empty} region is a false positive). 
Moreover, neural networks have a \textit{hard time} detecting small objects~\cite{cai2018cascade}. However, some of the smaller objects are captured at an even higher resolution than that used by the network's input layer, but once the captured image is resized to feed the neural network, these objects become just a few indistinguishable pixels. For example, an 8MP camera (4K UHD resolution, or 3840x2160 pixels) captures an object at 20 meters with a pixel density of over 100 px/m, enough to get a sharp image of a license plate. Nonetheless, if the image gets resized to standard definition (640x480 pixels), density drops below 20 px/m (calculated for a camera with 3mm focal length placed 4m height), which is unsuited for most detection tasks. For reference, 320x320 pixels is a common input size for \textit{edge} detection models.

We make the following key observation: scenes captured by static cameras do not move; objects of interest do. We do not need to check the entire scene, just whatever is new in it. By focusing on footage from static cameras, we can extract the regions of the scene that contain \textit{objects of interest} (i.e., objects not yet classified) and filter the rest of the image out. For this task, we have multiple techniques at our disposal that we can apply to identify and extract such regions of interest, like motion detection or background subtraction. This step alone lets the object detection model analyze objects at a higher resolution and, potentially, increase its accuracy. At the same time, we can optimize the analysis of multiple scenes by merging regions from different cameras into a composite frame that can be passed to the model as a single input. Therefore, we can effectively exploit the intrinsic parallelism of neural networks and work on multiple video streams simultaneously, reducing the total number of inferences required by the system. 

In this paper, we present FoMO (\textit{Focus on Moving Objects}). A method to optimize video analytics by removing the analysis of \textit{uninteresting} parts of the scene, distributing the work, and consolidating information from multiple cameras to reduce the compute requirements of the analysis. Moreover, as a by-product, neural networks can analyze objects from the scenes at a higher resolution, which results have shown to improve the accuracy of the models tested by several fold.

The paper is organized as follows: in Section~\ref{sec:background}, we introduce some of the concepts of computer vision and video analytics from which we base our work. Then, we present and describe FoMO in Section~\ref{sec:method}. In Section~\ref{sec:setup}, we detail the experimental setup and the methodology followed during the evaluation. In Section~\ref{sec:results}, we presents the results of FoMO's evaluation. The literature review and previous work is presented in Section~\ref{sec:related}. Finally, Section~\ref{sec:conclusions} concludes the paper.

%% file: background.tex
\section{Background and Considerations}
\label{sec:background}
In this section, we revise several considerations made in this paper and the concepts behind each one. 

\subsection{Video Analytics Systems}
Video analytics applications can be divided into \textit{online} and \textit{offline} video analytics, depending on \textit{when} the information extracted from the frames is needed. On the one hand, online video analytics requires images to be processed in real-time, as actions are triggered by the analysis of what is in front of the camera when images are captured. An ALPR (Automatic License Plate Recognition) system controlling a barrier to entry a given facility would fall under this category, as well as an autonomous car detecting obstacles to avoid them. At the same time, past information is no longer useful for online systems. Consequently, such systems are subject to strict latency constraints and are typically evaluated by the single inference latency. Therefore, they are unable to fully exploit parallelism due to the \textit{hazards} of request consolidation under strict deadlines.

On the other hand, offline video analytics processes images captured in the past. For example, a system to query the contents of recorded footage to obtain the fragments or timestamps that contain the queried objects~\cite{kang2017noscope} falls under this category. These systems analyze images in bulk and are commonly evaluated based on the turnaround latency of the queries (i.e., total system throughput). Therefore, they can prioritize resource utilization over single inference latency. 

Online and offline video analytics are two categories with very distinct requirements. In this paper, we focus on online video analytics for two reasons. First, they show subpar scalability as the set of available optimizations while scheduling requests or orchestrating available resources is limited. Second, offline video analytics can also benefit from an online analysis to pre-process images and reduce the search space of queries~\cite{hsieh2018focus}.

\subsection{Selection Region of Interest}
Object detection is, in fact, a combination of region proposal and image classification problems. Region proposal techniques suggest a set of regions in the input image that might contain objects of interest (i.e., objects the neural network was trained to detect). These regions are then processed as independent images and are classified based on their contents. The region is considered a positive example if the highest-scoring classification is above a user-defined threshold. Intuitively, the quality of the region proposal mechanisms has a high impact on the quality of the detection results. There are several methods available, such as Selective Search or Focal Pyramid Networks (FPN)\cite{lin2017feature}. 

Neural networks that differentiate region proposal and classification in two stages are called \textit{Two-Shot} detection models. On the contrary, \textit{Single-Shot} detection models skip the region proposal stage and yield localization and class predictions all at once. Regardless of the number of shots, all object detection models predict bounding boxes and classes using a single input layer of a fixed size decided during training. Larger input layers potentially yield more accurate predictions, as more pixels (i.e., information) are taken into account. This is especially true for smaller objects that are more difficult to detect correctly. However, any increment in the input layer's size is followed by an increment in the compute requirements that is not always matched with an equal increment in accuracy.

\subsection{Moving/Foreground/Salient Object Detection}
FoMO relies on background subtraction (BGS) methods that can detect and extract objects of interest. Such methods allow us to differentiate between \textit{background} and \textit{foreground} objects. However, which objects compose the background and foreground is not always clear and is subject to interpretation. For example, in most scenarios, a car would be considered part of the foreground, as cars come and go from the field of view of the camera and are often an object of interest to identify or detect (e.g., autonomous driving or smart cities). However, a broken car in a scrapyard will probably remain where it is for as long as other objects that are typically considered background objects (e.g., traffic lights or buildings). Once that car is correctly identified, there is no need to process it on successive frames repeatedly. Therefore, we make the following assumption: \textit{foreground} objects move, \textit{background} objects are stationary.

As for the set of techniques that can leverage the extraction of foreground objects, we have mainly considered traditional computer vision techniques that deliver \textit{relatively} good results at a small compute overhead. Newer and more accurate methods make use of neural networks to provide high-quality results~\cite{zeng2019combining}, but their compute requirements are higher than that of many neural networks, which significantly reduces the room for improving the system performance. Moreover, it is important to note that, while an accurate algorithm to detect salient objects is desirable, background subtraction methods that offer more modest results \textit{usually} increase the number of false positives instead of false negatives. That is, more or larger regions are flagged as foreground/moving than those that actually changed, which can be caused due to camera jitter, changes in the lighting conditions, or other phenomenons causing unwanted frame-to-frame variations. However, false positives increase the amount of work but do not necessarily translate into false positives being predicted by the detection model. Ultimately, the selection of a more or less accurate BGS method will depend on the compute capabilities available at the edge locations (cameras or compute next to the cameras).

%% file: method.tex
\section{Focus on Moving Objects}
\label{sec:method}

In this paper, we present FoMO (Focus on Moving Objects), a novel method to optimize and distribute video analytics workloads in Edge locations. FoMO has been conceived with static cameras in mind, and its goal is to maximize the \textit{pixel-to-object ratio} that is processed at each inference, i.e., it aims to maximize the number of pixels processed by the neural network that belong to actual objects of interest instead of the background. Towards this goal, each frame is preprocessed to extract the regions of interest that have a higher chance of containing objects of interest. Working with static cameras, we can safely assume that such regions of interest intersect with regions whose content has changed over time. 

\begin{figure*}
    \centering
    \includegraphics[width=\textwidth]{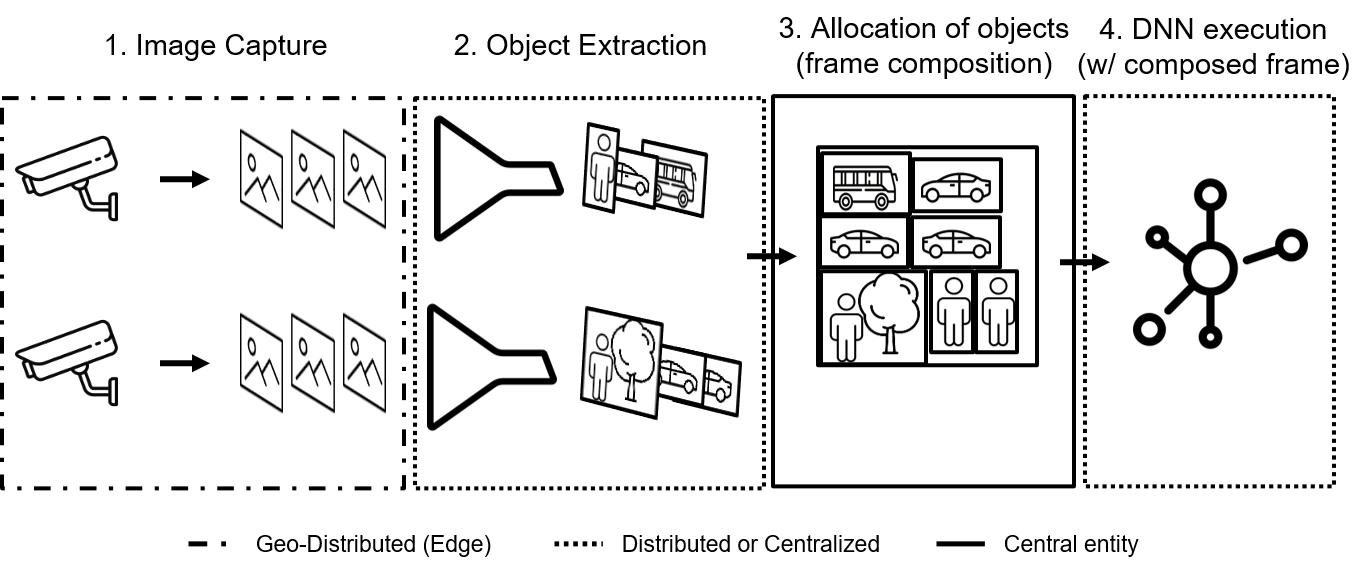}
    \caption{Main steps involved in the process of \methodname.}
    \label{fig:key_steps}
\end{figure*}

Figure \ref{fig:key_steps} depicts the main steps involved in FoMO. First, a set of static cameras (either in a single location or geographically distributed) periodically capture images from the scenes. Then, each scene is processed individually, and background subtraction is computed to extract moving objects (i.e., regions of the image where movement has been detected). Depending on the compute and network bandwidth available at the Edge, video scenes can be processed locally or sent to a central location. In both cases, a single entity, known as \textit{composer}, is in charge of receiving the objects from all the scenes and consolidating objects into a single RGB matrix known as \textit{composite frame}. This step is called \textit{frame composition}.

During frame composition, each object is treated as an independent unit of work that can be allocated separately or jointly from other objects from the same or other scenes. Objects are selected based on a pre-defined composition policy that determines the order and rate at which objects are composed and processed. The selected objects are placed together at each composition interval, forming a mosaic of objects (i.e., the composite frame). A few-pixels wide border is added to mark frontiers between objects. Next, the composite frame is used as input for the object detection model. Finally, the predicted coordinates of the detected objects are translated back to the original coordinates of the corresponding scene.

\subsection{Extraction of Objects of Interest}
Frame decoding results in a 3-D matrix containing the RGB color of every pixel in the scene captured by the camera. When cameras capturing the scene are static, consecutive RGB matrices will often have little to no pixel-to-pixel variation. At a higher level of abstraction, this correlation among frames means that nothing is moving in the scene (or the camera lens could not capture it). Consequently, whenever an object is moving in the scene, only some subregions are affected. This is the basis for \textit{motion detection} algorithms, which are extensively used to trigger the analysis of a given frame to save computation on frames that do not contain anything new. In this paper, however, we extend this idea and consider that moving objects, not whole frames, are of interest. The rationale is that for an object to trigger an action, it has to either enter, leave or interact with the scene (i.e., moving in the scene). Thus, we can focus on the regions that contain change and discard the rest. As these subregions are usually a tiny fraction of the whole frame, we can combine several subregions from different cameras into a composed frame. This frame can be processed by a neural network the same way it would process an entire frame.

Motion detection involves many challenges (e.g., camera jitter or variations in lighting, among others), and there is no single nor standard method that can handle all of them robustly. Moreover, it usually relies on background initialization algorithms to model the background without foreground objects. Among all available methods, we prioritized simplicity (i.e., speed) over accuracy for two reasons. First, the method must run in resource-constrained (Edge) nodes and do it faster than it would take to process the whole frame by the neural network. Second, the accuracy of these methods is usually evaluated by error metrics that compare the modeled background to the ground-truth background at the pixel level. However, artifacts and other inconsistencies in the generated background do not necessarily translate into a lower recall of objects of interest or a lower accuracy of the object detection model.

We have evaluated FoMO using three different BGS methods to extract objects of interest from a scene. These methods differ by the way they model the background and, second, by how they identify foreground objects. The three BGS methods are:
\begin{itemize}
    \item \textbf{PtP Mean} (Moving Average Pixel-to-Pixel): The background corresponds to the moving average of \textit{n} frames (not necessarily consecutive) at pixel level. Objects are extracted by applying first a Gaussian blur to the background model and the current frame to reduce noise. Then, the absolute difference between both frames is computed before a binary threshold decides which level of difference constitutes change and which does not. 
    \item \textbf{MOG2} (Mixture of Gaussians): Each pixel is modeled as a distribution over several Gaussians, instead of being a single RGB color. This method is better than the moving average at preserving the edges and is already implemented in OpenCV. 
    \item \textbf{Hybrid}: The background is modeled as a Mixture of Gaussians (MOG2), but objects are extracted applying the same operations as in the PtP Mean to detect differences between the background and the current frame.
\end{itemize}

As output, all three methods provide a black and white mask of the same dimensions as the frames. From the mask, the bounding boxes of the objects moving are easily extracted by detecting adjacent pixels. Bounding boxes of an area smaller than a pre-defined minimum are discarded. This threshold helps to filter minor variations out. However, it also defines the method's sensitivity, and the optimal value will vary from scene to scene depending on the minimum size of objects (in pixels) to detect. 

All three methods can run in the order of milliseconds in resource-constrained nodes. Performance and accuracy of these methods are analyzed in Section~\ref{sec:results}, where we explore their impact on the final results. In short, a more sensitive motion detector will generate false positives (i.e., non-interesting objects being extracted, like a tree with moving leaves) and, therefore, introduce more or larger regions than necessary to be composed in the following step. 

Effectively, the breakdown of the scene into objects opens the door to two major optimizations. On the one hand, we can decide which regions should be processed and which can be omitted. Potentially, this reduces the amount of data moved around and processed by the detection model. Consequently, it increases the pixel-to-object ratio, as we achieve the same results while processing only those pixels that we have considered of interest. On the other hand, video analytics can now be distributed while consuming less network bandwidth, as, potentially, only a fraction of the scene must be sent over the network.

\subsection{Composition Policies}
After moving objects have been extracted from the frames, the composer has a pool of them in the form of cropped images, each with a different size and aspect ratio. At each composition interval, the composer decides what objects are to be allocated in the resulting composite frame. Currently, objects are selected in a purely first-come, first-serve manner. However, not all objects can be allocated during the subsequent allocation cycle, as there is a limit to the number of objects composed in one frame. The threshold is not a fixed value and is set upon one of two pre-defined composition policies. 

To the composer, objects become the minimum unit of work to be allocated into composite frames. Therefore, we could argue that the composer treats composite frames as a type of resource, a resource that can be shared among requests (i.e., objects). At the same time, the composite frame can be considered elastic, as it can be overprovisioned by simply adding more objects to it. Nonetheless, as in any other type of shared resource, there is a point at which a higher degree of overprovisioning degrades the overall performance. Object detection models can locate and classify multiple objects within an input image and do it with a single forward pass over the network. However, there is a minimum number of pixels required for the model to successfully detect an object (closely related to the focal view and input size of the network). Similarly, the larger an object is, the better can a model detect it accurately.

During composition, the composer and the composition policy must consider the trade-off between the resource savings from consolidating more objects into a single composite frame and the accuracy drop involved. After a frame is composed, the resulting frame is resized to match the neural network's input size regardless of its original size. At the same time, adding one more object to the composition could potentially result in a larger frame, depending on whether the new object can be allocated without increasing the dimensions of the composite frame with all previous objects. The larger the resulting composite frame, the smaller its objects become after the frame is resized to feed the neural network. Consequently, the more objects allocated in a composite frame, the higher the amount of computation saved (i.e., fewer inferences per camera), but also the higher the impact on the accuracy of the model. Unfortunately, there is no known mechanism to determine beforehand where the limit is or even where the sweet spot is. 

FoMO implements two policies that limit the number of objects allocated in a single composite frame, albeit the exact number differs from composition to composition. The first policy, namely \textit{downscale limit policy}, limits the downscaling factor to which objects will be subjected after the composite frame is resized to feed the neural network's input. This is equivalent to setting an upper limit to the dimensions of the resulting composite frame. This policy prioritizes the quality of the detection results over the system's performance. The second policy, namely \textit{elastic policy}, does not set a hard limit but a limit on the number of camera frames to consider on each composition. That is, the dimensions of the composite frame can be arbitrarily large but the objects allocated belong to, at most, \textit{n} camera frames, being \textit{n} user-defined. Effectively, this policy prioritizes system's performance over quality of the detection results.

\subsection{Allocation Heuristic}
Composing objects into a composite frame can be seen as allocating 2D images into a larger 2D canvas. For the sake of simplicity, we consider objects to be 2-D during allocation, as the third dimension is always of size 3 in RGB images. The allocation can be mapped to the \textit{bin packing problem}. Bin packing is an optimization problem that tries to pack (allocate) items (objects) of different volumes into bins of a fixed volume (composite frame) while minimizing the number of bins used (blank spaces in the composite node). This problem is, unfortunately, known to be an NP-hard problem. Therefore, FoMO approximates a solution using a \textit{first-fit} heuristic, where objects are first sorted into decreasing width (arbitrary) order. For this task, a sub-optimal solution results in a composite frame with more blank spaces than needed. Blank spaces reduce the density of meaningful pixels (i.e., pixels that are part of an object of interest) and, ultimately, reduce the number of objects that a single composed frame can fit.

%% file: setup.tex
\section{Experimental Setup}
\label{sec:setup}

In this section, we provide the specific experimental setup used throughout the evaluation presented in Section~\ref{sec:results}.

\subsection{Dataset}
\label{sec:dataset}

For the evaluation, we have used the VIRAT dataset~\cite{oh2011large}. VIRAT contains footage captured with static cameras from 11 different outdoor scenes.

\subsubsection{Dataset Curation}
VIRAT contains frame-by-frame annotations for all objects in the scene, with bounding box and label (classes person, car, vehicle, bike, and object). However, annotations include both static and moving objects. We evaluate FoMO by its capability to detect moving objects. Therefore, we have curated the dataset to remove all annotations from static objects and avoid these objects artificially lowering the final accuracy. Thus, we consider objects static if their coordinates in a given frame do not change to 10 frames prior (10 has been arbitrarily chosen and corresponds to the frame skipping we used during the evaluation). Nonetheless, cameras often suffer from a slight jitter, mainly due to wind. Whenever that happens, the bounding boxes of a given object on consecutive frames may not perfectly match, even if the object has not moved. To avoid false positives in such cases, we still consider an object static when its coordinates remained static for at least 90\% of the frames.

\begin{figure}[h!]
    \centering
    \includegraphics[width=\linewidth]{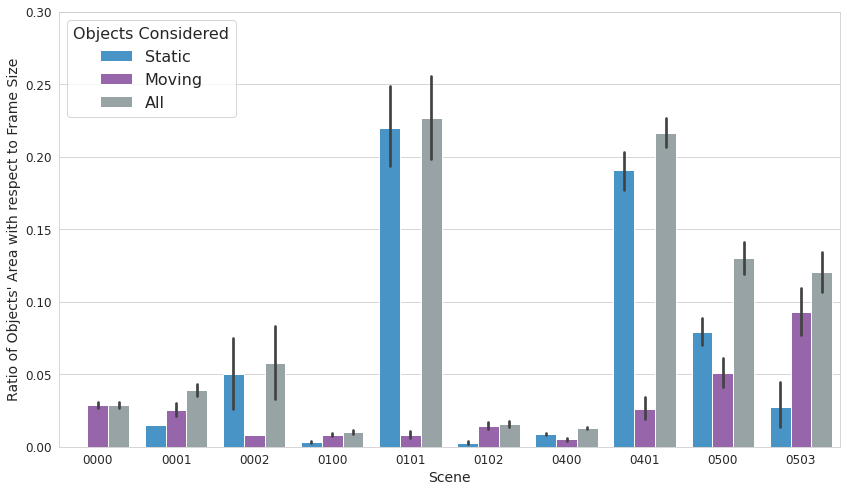}
    \caption{Percentage of area that represent objects in each scene, according to ground-truth annotations.}
    \label{fig:area_objects}
\end{figure} 

Figure~\ref{fig:area_objects} shows the percentage of area occupied by objects in each scene of the VIRAT dataset. Results show that scenes are mostly \textit{empty} and objects represent as little as 2\% of the scene and no more than a quarter of the scene. Moreover, some scenes appear to be \textit{quiet}, i.e., only a tiny fraction of objects in the scene are moving. When considering moving objects only, the average area taken by these objects fluctuates between 1\% and 8\% of the scenes. These results highlight the potential benefits that can be achieved by removing all the empty regions.

Moreover, we have made the following considerations during the evaluation:
\begin{itemize}
    \item Only sequences with at least 1000 frames have been considered.
    \item Discarded the first 250 frames (approx. 10 seconds) to give a time window to initialize the background modeling of some methods. 
    \item Frame skipping of 10 frames.
\end{itemize}

\subsection{Object Detection Models}
All experiments have been carried out using pre-trained object detection models that are publicly and freely available in the TensorFlow Model Zoo~\cite{tfmodelzoo}. We decided upon the use of pre-trained models instead of training or fine-tuning the models on our data to remove training as a variable on which to focus, which is an important one nonetheless.

The following are the three DNN models used throughout the evaluation:  
\begin{itemize}
    \item SSD + MobileNet V2 with an input layer of 320x320x3 pixels.
    \item Same as previous with an additional FPN (Feature Pyramid Network) that improves the quality of predictions~\cite{lin2017feature}.
    \item Same as previous with an input layer of 640x640x3 pixels to improve the quality of predictions.
\end{itemize}

\subsection{Background Subtraction Mechanisms}
Background subtraction (BGS) algorithms, albeit not directly part of our contributions, are key to extracting the set of objects that will constitute the system's workload. At the same time, they have an undeniable impact on the quality of the region proposal. An accurate BGS algorithm will extract all or most objects of interest and nothing else (i.e., discard all regions that do not contain objects of interest). However, to have a complete overview of how FoMO performs, we must break the accuracy of these methods down into \textit{precision} and \textit{recall}, as each one will have a different impact on the overall performance.

\textit{Precision} is defined as the ratio of \textit{True Positives} (TP) with respect to the sum of TP and \textit{False Positives} (FP), i.e., what proportion of predictions are correct. In the context of \textit{object extraction}, we can see precision at the pixel level and define it as the proportion of extracted area (i.e., number of pixels) that does belong to objects of interest with respect to the total extracted area (including that without objects of interest).
\textit{Recall} is defined as the ratio of TP with respect to the total number of positive examples (TP + FN), i.e. what proportion of positive examples are correctly detected. In the context of \textit{object extraction}, we define recall as the proportion of extracted area that does belongs to objects of interest with respect to the total area of the objects of interest in the scene (including the area of the objects not extracted).
Thus, low precision BGS mechanisms will cause \textit{uninteresting} regions to occupy space in the composed frame, potentially shrinking other regions or leaving them out of the composition (hence, delaying their processing). Low recall mechanisms will cause FoMO to directly miss predictions as entire objects were not extracted and will, therefore, not make it into the neural network's input. On the contrary, a high precision and high recall BGS mechanism will maximize FoMO's efficiency, while accuracy will be made to only depend on the neural network model chosen.

The experiments aim to quantify the impact that different methods for BGS, with more or less accuracy on the extraction of objects, have on the system's resource usage and the quality of the detections.

The configuration of the three BGS methods is, for all experiments, as follows:
\begin{itemize}
    \item PtP Mean: The moving average is computed over the last 20 frames with a frame skipping of 10 (i.e., spanning over 200 consecutive frames in total).
    \item MOG: Computed every frame. Implementation from OpenCV.
    \item Hybrid: MOG background model is updated once every 50 frames (2 seconds). The difference with the current frame to extract objects is computed with respect to the latest background model available.
\end{itemize}

\subsection{Metrics of Interest}
In video analytics, the two main metrics of interest are inference accuracy and cost. For the cost, we use latency as a proxy. For the inference accuracy, we use the average precision as used in the PASCAL Visual Object Classes (VOC) Challenge~\cite{everingham2010pascal}. To consider a detection as either positive or negative, we use a value of 0.3 for the Intersection over Union (IoU) between the predicted and the ground-truth bounding boxes. The IoU is a matric that measures the overlap between two bounding boxes.

%% file: results.tex
\section{Results}
\label{sec:results}
The following experiments evaluate the potential of our method by analyzing upper and lower-bound scenarios. Therefore, we have replicated the same stream whenever more than one stream is used unless otherwise stated. By replicating the stream, we avoid potential variability introduced by different load distributions across streams that is difficult to quantify. Consequently, if one object is captured at the \textit{i-th} frame of a stream, the same object will be captured in the same \textit{i-th} frame in all other streams. However, each frame's decoding and preprocessing (i.e., background subtraction, motion detection, and cropping) is computed for each stream individually to obtain an accurate and realistic performance evaluation.

\subsection{Accuracy Boost vs Resource Savings}
\label{sec:results-map}
\begin{figure}[h!]
    \centering
    \includegraphics[width=\linewidth]{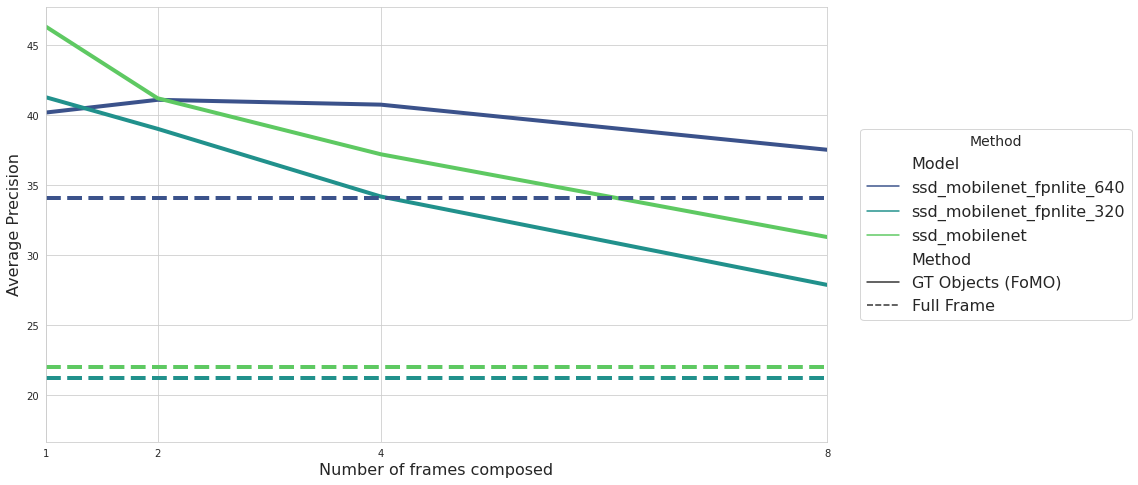}
    \caption{Mean Average Precision (mAP) by scene for three different DNNs with an increasing number of frames composed into a single inference. The reduction in number of inferences is equivalent to the number of frames composed.}
    \label{fig:overview_ap}
\end{figure}

Figure~\ref{fig:overview_ap} shows the mAP averaged for all scenes of the dataset for the three different pre-trained models. Each model is evaluated using the full frame as input compared to using a composed frame with objects from 1, 2, 4, and 8 parallel streams. The objects in each frame are extracted based on the ground truth annotations from the curated dataset, i.e., without background subtraction. Therefore, these results represent an upper bound precision for FoMO. It is important to understand why the mean average precision appears especially low for the baseline method (\textit{Full Frame}) during inference for the two smaller models. As mentioned in Section~\ref{sec:setup} (and, more specifically, as shown in Figure~\ref{fig:area_objects}), moving objects are, on average, a tiny fraction of the frame's area. Hence, once the frame gets downscaled to the neural network's input size, objects become too small to be detected accurately. This is why the bigger model, with an input twice as large as the other two models (640x640 pixels compared to 320x320 pixels), gets around 50\% improvement in accuracy over them.

From the mAP results of the models using FoMO, we extract two key observations. On the one hand, results show that a larger model does not necessarily translate into a higher mAP when inferencing over composite frames, as it happens with the baseline. The larger model is up to 10\% below the smaller models when the composite frame contains objects from only one camera frame. When the number of objects to compose is small (as it is expected when only a single camera frame is considered), the composition may result in a composite frame smaller than the model's input size. Consequently, the frame must be enlarged to the appropriate dimensions expected by the neural network. Thus, these results indicate that accuracy is also impacted negatively when objects are zoomed in. On the other hand, as already discussed, accuracy takes a hit when more frames (hence, more objects too) are considered during the composition, as objects become smaller. However, results show that a larger input size seems to mitigate this impact. The largest model still achieves 10\% higher accuracy composing frames eight cameras compared to the full-frame inference using the same model.

\subsection{Impact of BGS Techniques}
\label{sec:bgs_impact}

There are multiple available techniques to compute background subtraction to detect and extract moving objects in a scene. Each technique provides a different trade-off in terms of latency (or required resources) and quality of the solution. The quality of these techniques, however, can be measured through different metrics. We are interested in detecting moving objects but not \textit{everything} that is moving (e.g., leaves or camera jitter). Some methods are more sensitive to minor variations than others, and the level of sensitivity will impact what is considered actual movement and what is just background noise. Therefore, a higher sensitivity potentially results in a higher rate of false positives (i.e., lower precision) and a lower rate of false negatives (i.e., higher recall), as more objects will get extracted.

\begin{figure}[h!]
    \centering
    \includegraphics[width=\linewidth]{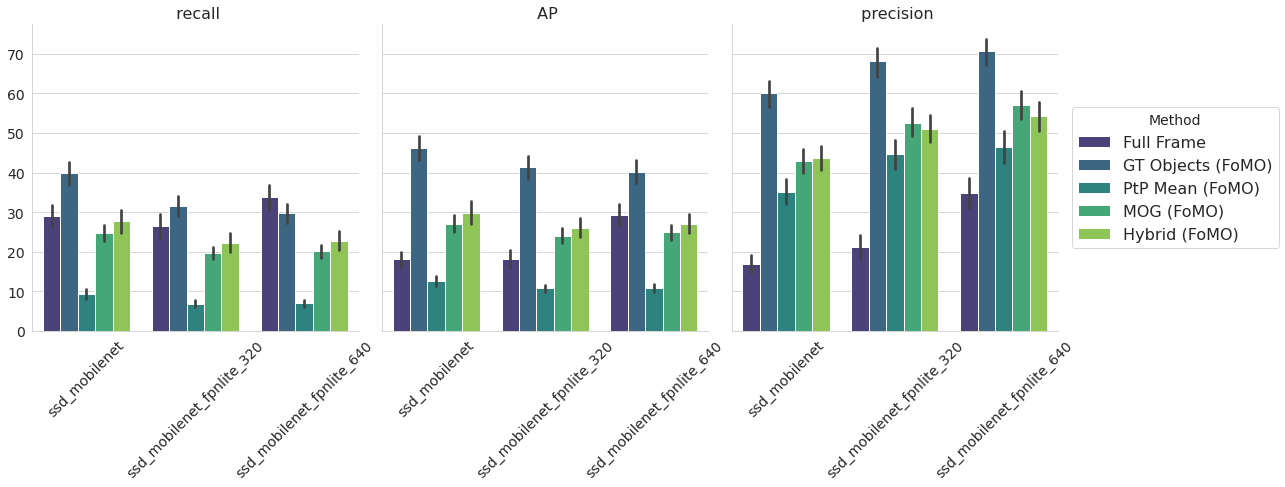}
    \caption{Recall, Precision, and Average Precision of three SSD MobileNet V2 pre-trained models after using the different methods to create the input for the model. \textit{Full Frame} uses the frame as captured by the camera; GT Objects: objects are cropped from the ground truth annotations; MOG: Mixture of Gaussians; PtP Mean: Pixel-to-Pixel Mean; Hybrid: MOG for background modeling and then pixel-to-pixel difference with respect the current frame.}
    \label{fig:bgs_map}
\end{figure}

Figure~\ref{fig:bgs_map} shows the recall, precision, and average precision of the three object detection models using the four BGS methods considered plus the baseline (\textit{Full Frame}, i.e., no object extraction). On the one hand, results show how FoMO performs compared to the baseline method. The upper bound for FoMO can be defined by the ground-truth annotations (\textit{GT Objects}), as no object is missed during the extraction. GT Objects consistently outperforms the baseline by a considerable margin on every metric and model, except one case. For example, GT Objects achieves twice the AP of the baseline for the two smaller models, although this margin reduces to 30\% when the larger model processes the objects. However, baseline surpasses FoMO's upper bound recall by a 10\% when using the largest model. This seems to be related to what results in Section~\ref{sec:results-map} showed when composition considers objects from only one frame, i.e., the resulting composite frame is smaller than the neural network's input, and objects must be enlarged. Nonetheless, a baseline's precision half of FoMO's mitigates this issue.

On the other hand, results show what can be expected from using a more or less accurate BGS method. MOG and Hybrid are close on all metrics. Again, precision is where FoMO seems to provide larger benefits, especially for small input sizes. Both MOG and Hybrid outperform baseline's precision by a factor of 2 using the smaller models and close to 50\% for the larger model. However, all BGS methods seem to miss objects during extraction, and recall takes a hit. Nevertheless, the AP of both methods still outperforms the baseline by a 60\% and 30\% when using the two smaller models, while baseline outperforms the other two by 12\% in AP using the larger model. Finally, PtP Mean lacks far behind on the recall, which hits its average precision. Nevertheless, results show that FoMO consistently increases precision regardless of the BGS method, while the accuracy of the selected method will mainly impact its recall.

\subsection{Performance and Resource Usage}
FoMO's main goal is to improve the overall system performance. This is achieved by reducing the amount of computing required to process a single frame. After decoding, each frame undergoes three consecutive steps: 1. object extraction, 2. frame composition, 3. inference. We now evaluate the first two, as the third will be determined by the neural network used. 

Table~\ref{tab:bgs_latency} shows the average latency to extract objects of interest for each of the three methods considered in the experiments. These results were obtained using a single core on an Intel Xeon 4114. Pixel-to-Pixel mean is, on average, eight times slower than MOG2 and almost six times slower than the Hybrid method. The table does not show timings for the executions that use ground truth annotations as those are synthetic executions that do not require any background subtraction to get the objects' bounding boxes.

\begin{table}[h]
\centering
\begin{tabular}{cclll}
\textbf{Method} & \textbf{Latency (millisec)} &  &  &  \\ \cline{1-2}
PtP Mean        & 22.0         &  &  &  \\
MOG2            & 2.7         &  &  &  \\
Hybrid          & 3.8         &  &  & 
\end{tabular}
\caption{Average latency (milliseconds) to extract objects using the three methods considered in the experiments}. 
\label{tab:bgs_latency}
\end{table}

\begin{figure}[h!]
    \centering 
    \includegraphics[width=\linewidth]{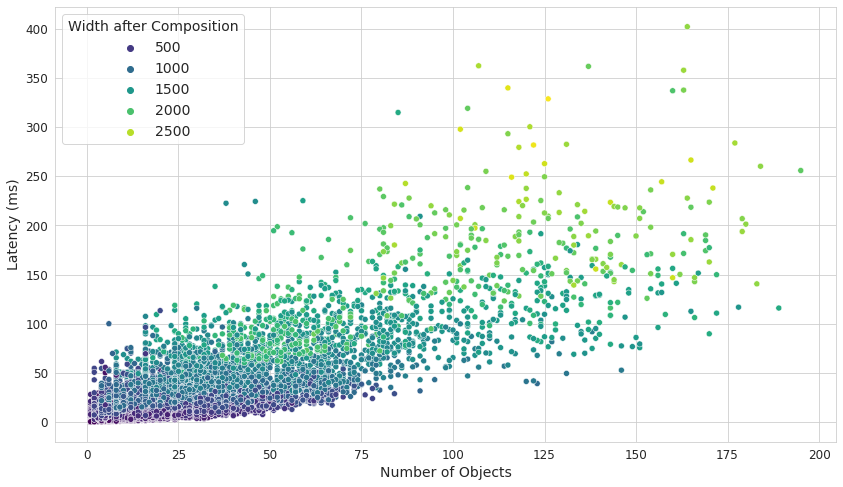}
    \caption{Time to compose a frame, i.e. latency, in milliseconds (Y-Axis) with respect to the number of objects to compose (X-Axis). The color palette shows the width of the resulting composite frame. Each dot is a composition with objects from up to eight parallel streams.}
    \label{fig:composition_latency}
\end{figure}

The next step is the composition of objects into the composite frame. Potentially, the composition is performed once for multiple scenes. Figure~\ref{fig:composition_latency} shows the time required to compose a frame with respect to the number of objects to compose and the width of the resulting composite frames (and height, as these are always 1:1). Results show how the latency increases with the number of objects to compose. The number of camera frames considered within a composition interval indirectly impacts the latency, as the more frames are considered, the more objects we can expect to be available for composition (although not necessarily always true). The increasing dimensions of the resulting composite frames highlight this relationship. The frame is enlarged to try to fit a larger number of objects. Nonetheless, we can expect the detection models to yield worse accuracy when processing the larger composite frames, as objects will appear smaller once shrunk to the model's input size. Therefore, datapoints with composition latency in hundreds of milliseconds are too large to be reliably used unless the detection model has an appropriately sized input layer. In that case, inference cost will also be higher, and, therefore, the cost of composing many objects can be hidden.

%% file: related_work.tex
\section{Related Work}
\label{sec:related}

Given the amount of data and compute required to train neural networks, the training step has garnered most of the attention from both academia and industry. However, as neural networks start being widely deployed for production use cases, optimizing inference is gaining interest from the industry. Most of the previous works has focused on the optimization of either the neural network architecture itself (using techniques to prune layers \cite{kang2017noscope} or reduce weight precision through quantization) or on the optimization of the application's pipeline (using cascade classifiers \cite{kang2017noscope,hsieh2018focus} or reducing the number of frames to process~\cite{canel2019scaling}). In the context of video analytics systems, there are two main directions: reducing the amount of work \cite{canel2019scaling} and optimizing the inference pipeline \cite{kang2017noscope}.

Regarding what impacts the quality of the predictions of a neural network, the work in~\cite{cai2018cascade} argues that, in general, a detector can only provide high quality predictions if presented with high quality [region] proposals. Similarly, authors in~\cite{eggert2017closer} demonstrate how deep neural networks have difficulties to accurately detect smaller objects by design.

Regarding the usage of traditional computer vision techniques, authors in 
Authors in~\cite{bouwmans2017scene} provide a taxonomy and evaluation (in terms of accuracy and performance) of scene background initialization algorithms, which helped us decide what were the most sensible methods to implement in FoMO.

Previous methods have focused on optimizing the different transformations applied to the input image. However, all these methods still devote a significant amount of computation to decide whether a certain region contains true positives or not (region proposal) when, in practice, most will not. Our method aims to reduce the amount of work wasted on this purpose. Moreover, this allows us to stack regions of different cameras to reduce the total computation of the system and increase the overall system throughput. At the same time, our results have shown that by increasing the quality of proposals, accuracy increases considerably as a \textit{side effect}.

To the best of our knowledge, no previous work proposes and evaluates a method combining traditional computer vision techniques to quantize and distribute portions of scene, consolidate multiple scenes into one, and process them with a single inference.

%% file: conclusions.tex
\section{Conclusions}
\label{sec:conclusions}
In this paper, we have presented FoMO, a method that distributes the load of video analytics and, at the same time, reduces the compute requirements of video analytics and improves the accuracy of the neural network models used for object detection. Results have shown how FoMO is able to scale the number of cameras being processed with a sub-linear increment in the amount of resources required while still improving accuracy with respect to the baseline. Results have shown that the number of inferences can be reduced by 8 while still achieving between 10\% and 40\% higher mean average precision with the same model. 
Moreover, the methodology used shows how video analytics can be effectively deployed in resource-constrained edge locations by tackling its optimization from a different perspective that opens a new line of research. 